\documentclass[10pt,twocolumn,letterpaper]{article}

\usepackage{cvpr}
\usepackage{times}
\usepackage{epsfig}
\usepackage{graphicx}
\usepackage{amsmath}
\usepackage{amssymb}
\usepackage{cite}
\usepackage{epstopdf}
\usepackage{booktabs}
\usepackage{subfigure}
\usepackage{multirow}
\usepackage{color}
\usepackage{authblk} 

\usepackage[breaklinks=true,bookmarks=false]{hyperref}

\cvprfinalcopy 

\def\cvprPaperID{****} 

\ifcvprfinal\fi
\begin{document}
\title{Self-critical $n$-step Training for Image Captioning}
\author[1]{Junlong Gao}
\author[4]{Shiqi Wang}
\author[2,3]{Shanshe Wang\thanks{Corresponding author}}
\author[2,3]{Siwei Ma}
\author[2,3]{Wen Gao}
\affil[ ]{$^{1}$Peking University Shenzhen Graduate School}
\affil[ ]{$^{2}$Institute of Digital Media, Peking University~~~~~$^{3}$Peng Cheng Laboratory}
\affil[ ]{$^{4}$Department of Computer Science, City University of Hong Kong, Hong Kong}

\maketitle
\thispagestyle{empty}

\begin{abstract}
	Existing methods for image captioning are usually trained by cross entropy loss, which leads to exposure bias and the inconsistency between the optimizing function and evaluation metrics. Recently it has been shown that these two issues can be addressed by incorporating techniques from reinforcement learning, where one of the popular techniques is the advantage actor-critic algorithm that calculates per-token advantage by estimating state value with a parametrized estimator at the cost of introducing estimation bias. In this paper, we estimate state value without using a parametrized value estimator. With the properties of image captioning, namely, the deterministic state transition function and the sparse reward, state value is equivalent to its preceding state-action value, and we reformulate advantage function by simply replacing the former with the latter. Moreover, the reformulated advantage is extended to $n$-step, which can generally increase the absolute value of the mean of reformulated advantage while lowering variance. Then two kinds of rollout are adopted to estimate state-action value, which we call self-critical $n$-step training. Empirically we find that our method can obtain better performance compared to the state-of-the-art methods that use the sequence level advantage and parametrized estimator respectively on the widely used MSCOCO benchmark.  
\end{abstract}

\section{Introduction}

Image captioning aims at generating natural captions automatically for images, which is of great significance in scene understanding. It is a very challenging task, which requires to recognize important objects in the image, as well as their attributes and relationships between each other, such that they can be finally described properly in natural language. The ability of the machine to mimic human in expressing rich information in natural language with correct grammar is important since it can be applied to human-robot interaction and blind users guiding.

Inspired by the recently introduced encoder-decoder framework for machine translation in \cite{Cho2014Learning}, most recent works in image captioning have adopted this paradigm to generate captions for images \cite{Vinyals2015Show}. In general, an encoder, \eg convolutional neural network (CNN), encode images to visual features, while a decoder, \eg long short term memory (LSTM) \cite{Hochreiter1997Long}, decodes the visual features to generate captions. These methods are trained in an end-to-end manner to minimize cross entropy loss, \ie maximize the likelihood of each ground-truth word given the preceding ground-truth word, which is also known as “Teacher Forcing”\cite{lamb2016professor}.

The first problem of cross entropy loss is that it will lead to “exposure bias”, since in the training stage, the model is only fed with ground-truth word at each time step, while in the testing stage, the model is fed with the previously predicted word. This discrepancy between training and testing easily results in error accumulation during generation, as the model is not exposed to its predictions during training and difficult to handle the errors which never occur in the training stage. In order to handle exposure bias, Bengio \etal \cite{bengio2015scheduled} feed back the model own predictions as input with scheduled sampling, while Lamb \etal \cite{lamb2016professor} proposed the “ Professor Forcing” on top of the “Teacher Forcing”.

The second problem of cross entropy loss is that the generated sentences are evaluated in the testing stage by non-differentiable metrics, such as BLEU 1-2-3-4 \cite{papineni2002bleu}, ROUGE \cite{lin2004rouge}, METEOR \cite{banerjee2005meteor}, CIDEr \cite{vedantam2015cider}, SPICE \cite{anderson2016spice}, while during training the model is trained to minimize cross entropy loss, which is the inconsistency between the optimizing function and evaluation metrics. The methods proposed in \cite{bengio2015scheduled,lamb2016professor} cannot address this inconsistency. Recently, it has shown that policy gradient algorithm in reinforcement learning (RL) can be trained to avoid exposure bias and directly optimize such non-differentiable evaluation metrics \cite{ranzato2015sequence,liu2017improved,zhang2017actor,rennie2017self}. In this way, the model can be exposed to its own predictions during training. However, the algorithms in \cite{rennie2017self} use sequence level advantage that implicitly makes an invalid assumption that every token makes the same contribution to the whole sequence. Many works\cite{ranzato2015sequence,liu2017improved,zhang2017actor} have been proposed to model per-token advantage. However, they utilize a parametrized value/baseline estimator at the cost of introducing estimation bias. 

In this paper, we improve the advantage actor-critic algorithm to estimate per-token advantage without introducing the biased parametrized value estimator. With the properties of image captioning, namely, the deterministic state transition function and the sparse reward, state value is equivalent to its preceding state-action value, and we reformulate advantage function by simply replacing the former with the latter. Since state-action value cannot be precisely estimated, the model may easily converge to the local maxima trained with the reformulated advantage function. Therefore, we propose $n$-step reformulated advantage function, which can generally increase the absolute value of the mean of reformulated advantage while lowering variance. In order to estimate state-action value, we use Monte Carlo rollouts inspired by \cite{liu2017improved,yu2017seqgan} and max-probability rollout inspired by \cite{rennie2017self}, which is termed as self-critical $n$-step training. According to the empirical results, our model improves the performance of image captioning compared to the methods that use the sequence level advantage and parametrized estimator respectively. 

Overall, we make the following contributions in this paper: (1) with the special properties of image captioning, we find the equivalence between state value and its preceding state-action value, and reformulate the original advantage function for each action; (2) on top of the reformulated advantage function, we extend to $n$-step reformulated advantage function to generally increase the the absolute value of the mean of reformulated advantage while lowering variance; (3) we utilize two kinds of rollout to estimate state-action value function to perform self-critical training.


\section{Related Work}

Many different models have been developed for image captioning, which can be divided into two categories: template-based methods \cite{farhadi2010every,kulkarni2013babytalk} and neural network-based methods. Since our method adopts neural network architecture, we mainly introduce methods in this vein. Efforts of this line have been devoted to two directions: attention mechanism and reinforcement learning.

\subsection{Attention Mechanism}

The encoder-decoder framework of machine translation \cite{Cho2014Learning} was firstly introduced by \cite{Vinyals2015Show}, which feeds the last fully connected feature of the image into RNN to generate the caption. Xu \etal \cite{xu2015show} proposed soft and hard attention mechanisms to model the human’s eye focusing on different regions in the image when generating different words. This work is further improved in \cite{anderson2017bottom,rennie2017self,lu2017knowing,chen2017sca}. In \cite{lu2017knowing}, they introduced a visual sentinel to allow the attention module to selectively attend to visual and language features. Anderson \etal \cite{anderson2017bottom} adopted a bottom-up module, that uses object detection to detect objects in the image, and a top-down module that utilizes soft attention to dynamically attend to these object features. Chen \etal \cite{chen2017sca} proposed a spatial and channel-wise attention model to attend to visual features. Rennie \etal \cite{rennie2017self} proposed FC model and Att2in models which achieve good performance. 

\subsection{Reinforcement Learning}

Recently a few works use reinforcement learning-based methods to address the exposure bias and the mismatch between the optimizing function and the non-differentiable evaluation metrics \cite{ranzato2015sequence,rennie2017self,liu2017improved,zhang2017actor} in image captioning. Ranzato \etal \cite{ranzato2015sequence} firstly introduced REINFORCE algorithm \cite{williams1992simple} to sequence training with RNNs. However, REINFORCE algorithm often results in large variance in gradient estimation. To lower the variance of the policy gradient, many works have introduced different kinds of baseline into REINFORCE algorithm. For example, the reward of the caption generated by the inference algorithm is adopted as the baseline in \cite{rennie2017self}, which uses sequence level advantage while the per-token advantage was not considered. A variety of algorithms proposed in \cite{ranzato2015sequence,liu2017improved,zhang2017actor} aim at modeling the per-token advantage. Ranzato \etal \cite{ranzato2015sequence} used a baseline reward parametric estimator. In \cite{liu2017improved}, they used FC layers to predict the baseline and used Monte Carlo rollouts to predict the state-action value function. In \cite{zhang2017actor}, they combined the advantage actor-critic algorithm and temporal difference learning, and used another RNN to predict the state value function. However, the value/baseline estimator was used in \cite{ranzato2015sequence,liu2017improved,zhang2017actor}, which introduces estimation bias. In this paper, we utilize the properties of image captioning to reformulate the advantage actor-critic method and use different kinds of rollout to estimate the state-action value function to calculate per-token advantage without introducing bias. 

\section{Methodology}
\subsection{Training with cross entropy loss}
Given an image $I$, the goal of image captioning is to generate a token sequence
$\mathbb{A} = \left\{ {{a_1},{a_2},...,{a_T}} \right\}$, ${a_t} \in A$, where $A$ is the dictionary. The captioning model predicts a token sequence starting with ${a_0}$ and ending with ${a_T}$, where ${a_0}$ is a special token BOS indicating the start of the sentence, and ${a_T}$ is also a special token EOS indicating the end of the sentence. In order to simplify the formulas, $T$ is denoted as the total length of a generated sequence, ignoring the fact that generated token sequences have different lengths. We use the standard encoder-decoder architecture for image captioning, where a CNN as an encoder, encodes an image $I$ to an image feature ${I_F}$, and a RNN can be adopted as a decoder to decode ${I_F}$ to output a token sequence $\mathbb{A}$. In this work, we adopt the Att2in model proposed by \cite{rennie2017self}. Given a ground-truth sequence $\left\{ {a_1^*,a_2^*,...,a_T^*} \right\}$, the model parameters $\theta $ are trained to minimize the cross entropy loss (XENT)
\begin{equation}
\small{L\left( \theta  \right) =  - \sum\limits_{t = 1}^T {\log \left( {{\pi _\theta }\left( {a_t^*\left| {a_{1:t - 1}^*,{I_F}} \right.} \right)} \right)}}
\label{eq1}
\end{equation}
where ${\pi _\theta }\left( {{a_t}\left| {{a_{1:t - 1}},{I_F}} \right.} \right)$ is a probability distribution of the token ${a_t}$ given the preceding generated tokens $\left\{ {{a_1},{a_2},...,{a_{t - 1}}} \right\}$ and the image feature ${I_F}$.

\subsection{Training using policy gradient}
\noindent\textbf{Problem formulation.}
To address both problems of the cross entropy loss described above, namely, the exposure bias and the inconsistency between the optimizing function and evaluation metrics, we incorporate the reinforcement learning into image captioning. Formally, we consider captioning process as a finite Markov process (MDP). Our captioning model introduced above can be viewed as an agent, which interacts with an environment (words and images). In the MDP setting $\left\{ {S,A,P,R,\gamma } \right\}$, $S$ is a state space, $A$ is an action space as well as the dictionary, $P\left( {{s_{t + 1}}|{s_t},{a_t}} \right)$ is state transition probability, $R({s_t},{a_t})$ is reward function and $\gamma  \in \left( {0,\left. 1 \right]} \right.$ is the discounted factor. The agent selects an action, that corresponds to generating a token, from a conditional probability distribution $\pi \left( {a\left| s \right.} \right)$ called policy. In policy gradient algorithms, we consider a set of candidate policies ${\pi _\theta }\left( {a\left| s \right.} \right)$ parametrized by $\theta$. The state ${s_t} \in S$ is considered as a list composing of the image feature ${I_F}$ and the tokens/actions $\left\{ {{a_0},{a_1},{a_2},...,{a_{t - 1}}} \right\}$ generated so far:
\begin{equation}
\small{{s_t} = \left\{ {{I_F},{a_0},{a_1},...,{a_{t - 1}}} \right\}}
\end{equation}
Here we define the initial state ${s_0} = \left\{ {{I_F}} \right\}$. At each time step, the RNN consumes ${s_t}$ and uses the hidden state of RNN to generate the next token ${a_t}$. With the definition of the state, we have the next state ${s_{t + 1}} = \left\{ {{s_t},{a_t}} \right\}$: we simply append the token ${a_t}$ to ${s_t}$. According to the process, the state transition function $P$ can be called deterministic state transition function. Formally, we have:
\begin{equation}
\small{P\left( {{s_{t + 1}}|{s_t},{a_t}} \right) \equiv 1}
\label{eq2}
\end{equation}

When the state ${s_t}$ is transferred to the next state  ${s_{t + 1}}$ by selecting action ${a_t}$, the agent receives reward ${r_t}$ issued from the environment. However, in image captioning, we can only obtain a reward $r = R({s_T},{a_T}) = R\left( {{a_{1:T}}} \right)$ when EOS token is generated and ${\left\{I_F,a_0 \right\}}$ is not considered in reward calculation. The reward $r$ is computed by evaluating the generated complete sentences compared with corresponding ground-truth sentences under an evaluation metric. Therefore, we define the reward for each action as follows:
\begin{equation}
\small{
	{r_t} = \left\{ \begin{array}{l}
	0,t < T\\
	r,t = T
	\end{array} \right.}
\label{eq3}
\end{equation}

In reinforcement learning, a value function is a prediction of the expected, accumulative, $\gamma$ discounted future reward, measuring how good each state, or state-action pair, is. We define the state-action value function ${Q^\pi }({s_t},{a_t})$ and the state value function ${V^\pi }({s_t})$ of the policy $\pi$ as follows:
\begin{equation}
\small{
	\begin{array}{l}
	{Q^\pi }({s_t},{a_t}) = {E_{{s_{t + 1}},{a_{t + 1}},... \sim \pi }}\left[ {\sum\limits_{l = 0}^T {{\gamma ^l}{r_{t + l}}} \left| {{S_t} = {s_t},{A_t} = {a_t}} \right.} \right]\\
	{V^\pi }({s_t}) = {E_{{a_t},{s_{t + 1}},... \sim \pi }}\left[ {\sum\limits_{l = 0}^T {{\gamma ^l}{r_{t + l}}} \left| {{S_t} = {s_t}} \right.} \right]
	\end{array}}
\label{eq4}
\end{equation}
where ${Q^\pi }({s_t},{a_t})$ is the expected $\gamma$ discounted accumulated reward under policy $\pi$ starting from taking action ${a_t}$ at state ${s_t}$, and ${V^\pi }({s_t})$ is the expected $\gamma$ discounted accumulated reward starting from state ${s_t}$. To simplify the notation, we denote ${E_{{a_t},{s_{t + 1}},... \sim \pi }}\left[  \cdot  \right]$ and ${E_{{s_{t + 1}},{a_{t + 1}},... \sim \pi }}\left[  \cdot  \right]$ with ${E_\pi }\left[  \cdot  \right]$ in the rest of paper. It is obvious that the difference between ${Q^\pi }({s_t},{a_t})$ and ${V^\pi }({s_t})$ lies in whether taking the action ${a_t}$ or not at state ${s_t}$ when calculating the accumulated reward. In reinforcement learning, the agent aims to maximize the circumulative reward $L\left( \theta  \right) = {V^\pi }({s_0}) = {E_\pi }\left[ {\sum\nolimits_{t = 1}^T {{\gamma ^{t - 1}}{r_t}} } \right]$ by estimating the gradient ${\nabla _\theta }L\left( \theta  \right)$ and updating its parameters, instead of minimizing the cross entropy loss as Eq.~(\ref{eq1}).

In policy gradient methods, the gradient ${\nabla _\theta }L\left( \theta  \right)$ can be written as:
\begin{equation}
\small{{\nabla _\theta }L\left( \theta  \right) = {E_\pi }\left[ {\left( {{Q^\pi }({s_t},{a_t}) - b({s_t})} \right){\nabla _\theta }\log {\pi _\theta }\left( {{a_t}|{s_t}} \right)} \right]}
\label{eq5}
\end{equation}
where the baseline $b({s_t})$ can be any arbitrary function, as long as it does not depend on action ${a_t}$. This baseline does not change the expected gradient, but can decrease the variance of the gradient estimate significantly. This algorithm is known as REINFORCE with a Baseline.
Using ${V^\pi }({s_t})$ as the baseline $ b({s_t})$, the algorithm is changed to advantage actor-critic (A2C) algorithm as follows:
\begin{equation}
\small{{\nabla _\theta }L\left( \theta  \right) = {E_\pi }\left[ {{A^\pi }({s_t},{a_t}){\nabla _\theta }\log {\pi _\theta }\left( {{a_t}|{s_t}} \right)} \right]}
\label{eq6}
\end{equation}
In Eq.~(\ref{eq6}), ${A^\pi }\left( {{s_t},{a_t}} \right) = {Q^\pi }({s_t},{a_t}) - {V^\pi }({s_t})$ is called advantage function. This equation intuitively guides the agent to an evolution direction that increases the probability of better-than-average actions and decrease the probability of worse-than-average actions \cite{zhang2017actor}.  

\begin{figure}[t]
	\begin{center}
		\includegraphics[scale=0.26]{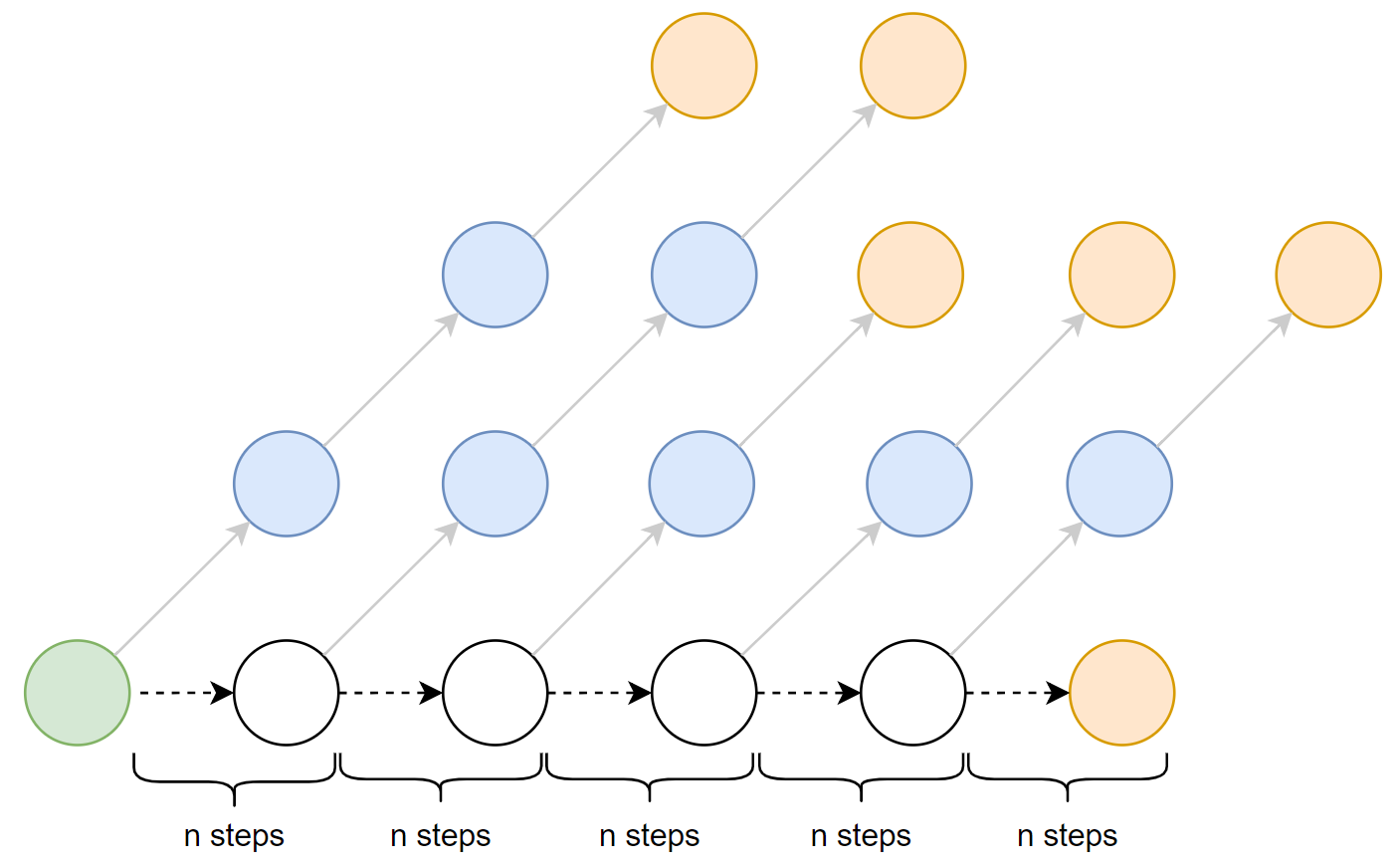}
	\end{center}
	\caption{Each state-action value is estimated by the average rewards of $K$ rollouts sequences ($K = 1$) or a reward of a max-probability rollout. The advantage function in our method is estimated by the current state-action value minus the preceding $n$-step state-action value. The tokens in green and yellow are the special tokens BOS and EOS. The tokens in white are the Monte Carlo trajectory and the tokens in blue are continuation rollout tokens for state-action value estimation. The $n$ steps in the figure means that the model performs rollouts every $n$ steps in $n$-step reformulated advantage function.}
	\label{fig1}
\end{figure}

\noindent\textbf{1-step reformulated advantage function.} 
Image captioning is a special case in reinforcement learning, for its state transition is deterministic, while other applications can have different next states with a certain probability, such as Atari Games. Here we use this property to reformulate Eq.~(\ref{eq6}).

With the definition of ${Q^\pi }({s_t},{a_t})$ and ${V^\pi }({s_t})$ in Eq.~(\ref{eq4}), we have
\begin{equation}
\small{{Q^\pi }\left( {{s_{t - 1}},{a_{t - 1}}} \right) = {r_{t - 1}} + \gamma \sum\limits_{{s_t} \in S} {P\left( {{s_t}\left| {{s_{t - 1}},{a_{t - 1}}} \right.} \right){V^\pi }({s_t})}}
\label{eq7}
\end{equation}
Due to the deterministic state transition function described above in Eq.~(\ref{eq2}), Eq.~(\ref{eq7}) can be rewritten as
\begin{equation}
\small{{{Q^\pi }\left( {{s_{t - 1}},{a_{t - 1}}} \right) = {r_{t - 1}} + \gamma {V^\pi }({s_t})}}
\end{equation}

In this paper, we set discounted factor $\gamma = 1$. According to reward function of Eq.~(\ref{eq3}), when $t \le T$, we have ${r_{t - 1}} = 0$. Then ${V^\pi }({s_t})$ can be written as 
\begin{equation}
\small{{V^\pi }({s_t}) = {Q^\pi }\left( {{s_{t - 1}},{a_{t - 1}}} \right)}
\label{eq8}
\end{equation}
Eq.~(\ref{eq8}) indicates that given the two properties of image captioning, namely the deterministic state transition function and the reward function, state value is equivalent to its preceding state-action value. Then we can rewrite Eq.~(\ref{eq6}) by incorporating Eq.~(\ref{eq8}) into Eq.~(\ref{eq6}) as follows:
\begin{equation}
\small{{\nabla _\theta }L\left( \theta  \right) = {E_\pi }\left[ {A_R^\pi ({s_t},{a_t}){\nabla _\theta }\log {\pi _\theta }\left( {{a_t}|{s_t}} \right)} \right]}
\label{eq9}
\end{equation}
where ${A_R^\pi }\left( {{s_t},{a_t}} \right) = {Q^\pi }({s_t},{a_t}) - {Q^\pi }({s_{t - 1}},{a_{t - 1}})$ is the refomulated advantage function from ${A^\pi }\left( {{s_t},{a_t}} \right)$ in Eq.~(\ref{eq6}). Therefore, ${Q^\pi }({s_{t - 1}},{a_{t - 1}})$ is a new baseline of ${Q^\pi }({s_t},{a_t})$ instead of ${V^\pi }({s_t})$. Each state-action value uses its preceding state-action value as baseline, such that it is termed as 1-step reformulated advantage function. 

In our approach, the agent aims at maximizing Eq.~(\ref{eq9}) rather than Eq.~(\ref{eq6}). Eq.~(\ref{eq9}) has an intuitive interpretation that it helps the agent to increase the probability of the action which has larger expected accumulated rewards compared to that of preceding action and decrease the probability of the action which has smaller expected accumulate rewards compared to that of preceding action. 

The most straightforward way to simulate the environment with the current policy $\pi$ is to obtain a Monte Carlo trajectory $\left\{ {\left( {{s_t},{a_t},{r_t}} \right)} \right\}_{t = 1}^T$ from the multinomial strategy and estimate the gradient ${\nabla _\theta }L\left( \theta  \right)$:
\begin{equation}
\small{{\hat \nabla _\theta }L\left( \theta  \right) = \frac{1}{T}\sum\limits_{t = 1}^T {\hat A_R^\pi ({s_t},{a_t}){\nabla _\theta }\log {\pi _\theta }\left( {{a_t}|{s_t}} \right)}}
\end{equation}
where ${\hat A_R^\pi }\left( {{s_t},{a_t}} \right) = {\hat Q^\pi }({s_t},{a_t}) - {\hat Q^\pi }({s_{t - 1}},{a_{t - 1}})$, and ${\hat Q^\pi }({s_t},{a_t})$ is an empirical estimate of ${Q^\pi }({s_t},{a_t})$.

\noindent\textbf{$n$-step reformulated advantage function.}
According to the property of Eq.~(\ref{eq9}) described above, the model encourages tokens better than its preceding token in terms of the value, and surpress the worse tokens. Though Eq.~(\ref{eq9}) is a greedy algorithm, Eq.~(\ref{eq9}) can guide the evolution direction of the model towards the global maxima only when state-action value is estimated precisely. Image captioning is considered as a model-free reinforcement learning task, which uses rollouts or function approximation to estimate state-action value. However, both methods, where the former suffers from a large variance and the latter introduces estimation bias, cannot predict absolutely precise value that may turn out to be wrong to encourage or suppress a token in this strict greedy strategy. Therefore, we introduce $n$-step reformulated advantage function. In $n$-step reformulated advantage function, we view $n$ steps as a large step to perform Eq.~(\ref{eq9}). Each step within the large step shares the $n$-step reformulated advantage $\hat A_R^\pi ({s_t},{a_t})$ as follows:
\begin{equation}
\small{\hat A_R^\pi \left( {{s_t},{a_t}} \right) = {{\hat Q}^\pi }({s_{\tau {\rm{ + }}n}},{a_{\tau {\rm{ + }}n}}) - {{\hat Q}^\pi }({s_\tau },{a_\tau })}
\label{eq9_5}
\end{equation}
where $\tau {\rm{ = }}\left\lfloor {t/n} \right\rfloor n$ and $\left\lfloor  \cdot  \right\rfloor$ is denoted as a round-down function, and $n$ ranges from $1$ to $T$ which unifies the two extremes, namely, $1$-step and $T$-step. In the $n$-step reformulated advantage, $n$ steps show a much clearer evolution trend of the Monte Carlo trajectory from the multinomial strategy than $1$ step, and the values of neighboring states in $n$-step have a more precise margin than that in $1$-step, except that state-action value estimation use the same strategy of Monte Carlo trajectory that samples one sequence from multinomial strategy. If they use the same strategy, estimated values of each time step are from the same distribution and thus larger $n$ cannot enlarge the margin of neighboring state values. Therefore, except for that particular case, as $n$ increases, the absolute value of the mean and the variance of reformulated advantage will be increased and reduced respectively. 

However, as $n$ increases, per-token advantage is inevitably gradually lost until a sequence level advantage of $n=T$. Therefore, different methods of estimating state-action value have different distributions and have different most suitable $n$ that performs best in balancing the approximation of per-token reformulated advantage and the improvement on the absolute mean of reformulated advantage. In general, the performance of small $n$ is better than that of $n=T$.

\noindent\textbf{Estimating the state-action value function.}
According to Eq.~(\ref{eq9_5}), we only need to estimate ${\hat Q^\pi }({s_t},{a_t})$. Here, we propose two methods to estimate non-parametric ${\hat Q^\pi }({s_t},{a_t})$: use $K$ Monte Carlo rollouts inspired by \cite{liu2017improved,yu2017seqgan} and use inference algorithm (max-probability rollout) inspired by \cite{rennie2017self}. These processes are illustrated as Fig.~\ref{fig1}. Since ${Q^\pi }({s_t},{a_t})$ is an expected accumulated reward, $K$ Monte Carlo is more stable and precise than max-probability rollout to estimate ${Q^\pi }({s_t},{a_t})$ with additional computation cost of $K-1$ rollouts. In 1-step reformulated advantage function, the model rollouts every steps, while in $n$-step reformulated advantage function, the model rollouts every $n$ steps. Therefore, the model adopts self-critical training \cite{rennie2017self}, which uses rollouts to estimate value functions as a critic.

In $K$ Monte Carlo rollouts, we sample $K$ continuations of the sequence $\left\{ {{s_t},{a_t}} \right\}$ to obtain $\left\{ {{a_{t + 1}},{a_{t + 2}},...,{a_T}} \right\}$, which means that the subsequent tokens are sampled from the multinomial strategy. When $\gamma  = 1$, according to Eq.~(\ref{eq3}) and Eq.~(\ref{eq4}),  the state-action value function can be computed by the average of the $K$ rewards 
\begin{equation}
\small{{\hat Q^\pi }({s_t},{a_t}) = \frac{1}{K}\sum\limits_{k = 1}^K {R\left( {{a_{1:t}};a_{t + 1:T}^k} \right)}}
\end{equation}
where ${R\left( {{a_{1:t}};a_{t + 1:T}^k} \right)}$ is denoted as the reward of the $k$’th continuation sampled after $\left\{ {{s_t},{a_t}} \right\}$ from the multinomial strategy. In our experiment, we set $K = 5$. A slight difference between our method and \cite{liu2017improved,yu2017seqgan} is that we need to rollout from $\left\{ {{s_0},{a_0}} \right\}$ to estimate ${Q^\pi }({s_0},{a_0})$, and they do not. If $K=1$, state-action value estimation and Monte Carlo trajectory both sample a sequence from multinomial strategy in each step, and thus larger $n$ cannot enlarge the margin of neighboring state values as discussed above. As $K$ increases, though $K$ rollouts of estimating state-action value are also sampled from multinomial strategy, the mean reward of $K$ can estimate more precise state-action value than $K=1$ (\ie state-action value estimation and Monte Carlo trajectory use different strategies in $K>1$) and thus larger $n$ will have larger absolute value of the mean of reformulated advantage with lower variance.

In max-probability rollout, we sample only one continuations of the sequence $\left\{ {{s_t},{a_t}} \right\}$ to obtain $\left\{ {{{\hat a}_{t + 1}},{{\hat a}_{t + 2}},...,{{\hat a}_T}} \right\}$, which are tokens of the largest probabilities at every time step. Then we have
\begin{equation}
\small{{\hat Q^\pi }({s_t},{a_t}) = R\left( {{a_{1:t}};{{\hat a}_{t + 1:T}}} \right)}
\end{equation}
where $R\left( {{a_{1:t}};{{\hat a}_{t + 1:T}}} \right)$ means the reward of the max-probability rollout sequence after $\left\{ {{s_t},{a_t}} \right\}$ under the inference algorithm. Interestingly, SCST \cite{rennie2017self} is equivalent to $T$-step reformulated advantage function using max-probability rollout, \ie SCST is a variant of ours. Here, state-action value estimation and Monte Carlo trajectory use different strategies, where the former are from max-probability strategy and the latter are from multinomial strategy. Moreover, max-probability strategy can always obtain better sequence than multinomial strategy. Therefore, though the reward of max-probability rollout cannot reflect the real state-action value, larger $n$ can have larger absolute value of the mean of reformulated advantage with lower variance.

It is worth noting that the rollout of preceding step can be used both in preceding token and this token with different effects. Here, we directly optimize CIDEr metric, \ie $R$ is CIDEr score. Moreover, only when calculating the last reformulated advantage of each sequence that includes token EOS, we use CIDEr with EOS as a token. Otherwise, we use CIDEr without EOS as a token. It is because EOS is not a normal token of a sentence like other words but a special token indicating the ending of the sentence, and it is ignored in the standard calculation of evaluation metric scores. 

\section{Experiments}

\subsection{Dataset}

We evaluate our method on the MSCOCO dataset \cite{lin2014microsoft}. For fair comparisons, we use the widely used splits from \cite{Karpathy2015Deep}. The training set contains $113,287$ images with $5$ captions for each image and $5K$ images for validation and $5K$ images for offline testing. We follow the standard practice to preprocess all captions, including converting all captions to lower case, tokenizing on white space, truncating captions longer than $16$ words, and replacing words that do not occur at least $5$ times with UNK token resulting in $9487$ words in the dictionary. To evaluate generated caption quality, we use the standard metrics, namely BLEU $1$-$2$-$3$-$4$, ROUGE, METEOR, CIDEr, SPICE. 
We extract image features using Resnet-101 \cite{he2016deep} without finetune.

\subsection{Implementation Details}
The embedding dimensions of the LSTM hidden, image, word and attention are all fixed to $512$ for all the models. We pretrain all the models under XENT loss for $30$ epochs using ADAM \cite{kingma2014adam} optimizer with default settings and fixed learning rate ${\rm{4}} \times {\rm{1}}{{\rm{0}}^{{\rm{ - 4}}}}$. During training under XENT loss, our batch size is set to $80$. We then run RL training with a fixed learning rate ${\rm{5}} \times {\rm{1}}{{\rm{0}}^{{\rm{ - 5}}}}$. In RL training, we use the models trained under XENT loss as the pretrained model in order to reduce the search space, and the batch size is set to $32$. In the whole training process, we use fixed dropout rate $0.5$ to prevent the models from overfitting.

\subsection{Experiment Configuration}
Here are the configurations of the basic model and several variants of our models. This series of experiments are designed to explore the effects of different $n$-step, different combinations of $n$ and $K$ Monte Carlo rollouts versus max-probability rollout. Besides, we re-implement two state-of-the-art reinforcement learning-based model SCST \cite{rennie2017self} and PG-CIDEr \cite{liu2017improved}, and all the hyperparameters are the same as those of our proposed models for fair comparison.

(1) XENT is the basic model trained with cross entropy loss, which is then used as the pretrained model of all reinforcement learning-based models.

(2) For max-probabilty rollout, we conduct $n$-step-maxpro ($n=1,2,4$) that are trained with $n$-step reformulated advantage throughout the whole training time. We also conduct models trained with different $n$-step successively, \eg $1$-$2$-$4$-$T$-step-maxpro, $T$-$4$-$2$-$1$-step-maxpro, $1$-$2$-$2$-step-maxpro.

(3) For $K$ Monte Carlo rollouts, we conduct $1$-step-sample that is trained with 1-step reformulated advantage using $K$ Monte Carlo rollout to estimate the state-action value function. We also conduct $1$-$2$-$2$-step-sample.

(4) SCST \cite{rennie2017self} (\ie $T$-step-maxpro) uses sequence level advantage for every token in a sampled sequence. Here, we compare self-critical per-token advantage with self-critical sequence level advantage. 

(5) PG-CIDEr \cite{liu2017improved} uses $K$ Monte Carlo rollouts with a parametrized estimator. Here, we compare self-critical per-token advantage with parametrized per-token advantage.

\begin{table*}
	\begin{center}
		{\small
			\begin{tabular}{lcccccccc}
				\toprule 
				
				& BLEU-1& BLEU-2& BLEU-3& BLEU-4& METEOR& ROUGE-L& CIDEr& SPICE\\
				\hline  
				XENT& 74.1& 57.4& 42.8& 31.7& 25.8& 54.1& 102.1& 19.2\\
				\hline
				PG-CIDER\cite{liu2017improved}& 77.44& 60.66& 45.85& 34.32& 26.35& 55.61& 113.9& 19.25\\
				SCST\cite{rennie2017self}($T$-step-maxpro)& 76.83& 60.65& 46.05& 34.61& 26.65& 56.03& 112.7& 19.99\\
				\hline
				$1$-step-sample& 77.49& 61.19& 46.64& \textbf{35.08}& 26.88& 56.11& \textbf{115.4}& 20.05\\
				$1$-$2$-$2$-step-sample& 77.41& 61.10& 46.46& 34.88& 26.88& 56.10& 114.9& 20.23\\
				\hline 
				$1$-step-maxpro& 77.24& 60.90& 46.13& 34.46& 26.87& 56.11& 115.1& 20.26\\ 
				$2$-step-maxpro& 77.82& 61.30& 46.45& 34.80& \textbf{26.95}& \textbf{56.29}& 114.6& 20.35\\
				$4$-step-maxpro& 77.67& 61.01& 46.30& 34.78& 26.91& 56.05& 114.5& 20.20\\ 
				$1$-$2$-$4$-$T$-step-maxpro& 77.45& 61.02& 46.25& 34.59& 26.89& 56.26& 114.8& 20.38\\
				$T$-$4$-$2$-$1$-step-maxpro& 77.30& 60.77& 46.07& 34.48& 26.74& 56.02& 114.0& 20.16\\
				$1$-$2$-$2$-step-maxpro& \textbf{77.93}& \textbf{61.54}& \textbf{46.75}& 34.96& 26.92& 56.27& 115.2& \textbf{20.42}\\
				\bottomrule 
			\end{tabular}
		}
	\end{center}
	\caption{Performance of our proposed models versus state-of-the-art models on the test portion of the Karpathy splits using greedy search.}
	\label{tabel1}
\end{table*}
\begin{table*}
	\begin{center}
		{\small
			\begin{tabular}{lcccccccccccccc}
				\toprule 
				\multicolumn{1}{c}{}& \multicolumn{2}{c}{BLEU-1}& \multicolumn{2}{c}{BLEU-2}& \multicolumn{2}{c}{BLEU-3}& \multicolumn{2}{c}{BLEU-4}& \multicolumn{2}{c}{METEOR}& \multicolumn{2}{c}{ROUGE-L}& \multicolumn{2}{c}{CIDEr}\\
				& c5& c40& c5& c40& c5& c40& c5& c40& c5& c40& c5& c40& c5& c40\\
				\hline  
				Google NIC\cite{Vinyals2015Show}&   71.3& 89.5& 54.2& 80.2& 40.7& 69.4& 30.9& 58.7& 25.4& 34.6& 53.0& 68.2& 94.3& 94.6\\
				Hard-Attention\cite{xu2015show}&    70.5& 88.1& 52.8& 77.9& 38.3& 65.8& 27.7& 53.7& 24.1& 32.2& 51.6& 65.4& 86.5& 86.3\\
				MSRCap\cite{fang2015captions}&      71.5& 90.7& 54.3& 81.9& 40.7& 71.0& 30.8& 60.1& 24.8& 33.9& 52.6& 68.0& 93.1& 93.7\\
				mRNN\cite{mao2014deep}&             71.6& 89.0& 54.5& 79.8& 40.4& 68.7& 29.9& 57.5& 24.2& 32.5& 52.1& 66.6& 91.7& 93.5\\
				ATT\cite{you2016image}&             73.1& 90.0& 56.5& 81.5& 42.4& 70.9& 31.6& 59.9& 25.0& 33.5& 53.5& 68.2& 94.3& 95.8\\
				Adaptive\cite{lu2017knowing}&       74.8& 92.0& 58.4& 84.5& 44.4& 74.4& 33.6& 63.7& 26.4& 35.9& 55.0& 70.5& 104.2& 105.9\\
				MIXER\cite{ranzato2015sequence}&    74.7&    -& 57.9&    -& 43.1&    -& 31.7&    -& 25.8&    -& 54.5&    -& 99.1&  -\\
				PG-SPIDEr\cite{liu2017improved}&    75.1& 91.6& 59.1& 84.2& 44.5& 73.8& 33.1& 62.4& 25.5& 33.9& 55.1& 69.4& 104.2& 107.1\\
				AC\cite{zhang2017actor}&    \textbf{77.8}& 92.9& 61.2& 85.5& 45.9& 74.5& 33.7& 62.5& 26.4& 33.4& 55.4& 69.1& 110.2& 112.1\\
				SCST-Att2in(Ens. 4)\cite{rennie2017self}&    -& -&  -& -&  -& -&  34.4& -& 26.8& -&  55.9& -&  112.3&  -\\
				\hline
				$1$-step-maxpro&                          77.1& 92.5& 60.6& 85.1& 45.8& 74.9& 34.1& 63.5& 26.6& 35.2& 55.6& 70.0& 111.1& 114.0\\
				$1$-step-sample&                          77.3& 92.5& 60.9& 85.4& 46.2& 75.2& 34.5& 64.0& 26.6& 35.2& 55.6& 70.2& 111.6& 114.5\\
				$1$-$2$-$2$-step-maxpro& 77.4& 92.9& 60.9& 85.6& 46.0& 75.2& 34.3& 63.7& 26.7& 35.2& 55.8& 70.0& 111.3& 113.5\\
				$1$-$2$-$2$-step-maxpro(Ens. 4)& 77.6& \textbf{93.1}& \textbf{61.3}& \textbf{86.1}& \textbf{46.5}& \textbf{76.0}& \textbf{34.8}& \textbf{64.6}& \textbf{26.9}& \textbf{35.4}& \textbf{56.1}& \textbf{70.4}& \textbf{112.6}& \textbf{115.3}\\
				\bottomrule 
			\end{tabular}
		}	
	\end{center}
	\caption{Leaderboard of published image captioning models on the official MSCOCO evaluation server.}
	\label{tabel2}
\end{table*}

\begin{figure*}[t]
	\begin{center}
		\begin{tabular}{cccc}
			\centering
			\subfigure[CIDEr]{
				\begin{minipage}[]{1.5in}
					\centering
					\includegraphics[height=0.95in, width=1.6in]{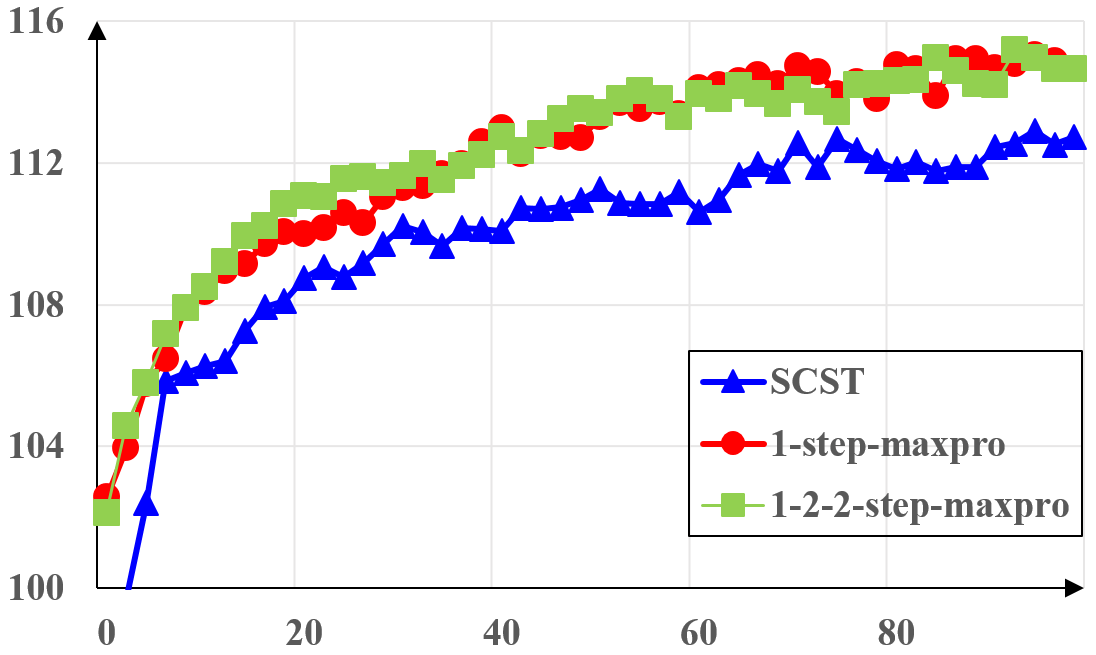}
				\end{minipage}	
				\label{fig2:a}
			}&
			\subfigure[METEOR]{
				\begin{minipage}[]{1.5in}
					\centering
					\includegraphics[height=0.95in, width=1.6in]{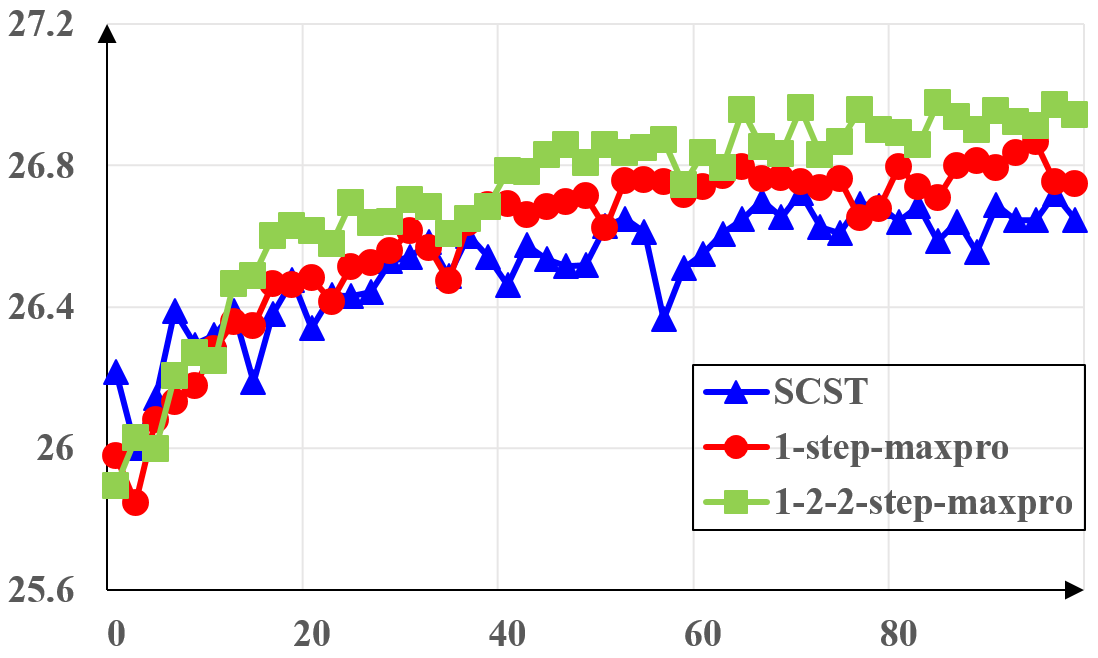}
				\end{minipage}
				\label{fig2:b}
			}&
			\subfigure[CIDEr]{
				\begin{minipage}[]{1.5in}
					\centering
					\includegraphics[height=0.95in, width=1.6in]{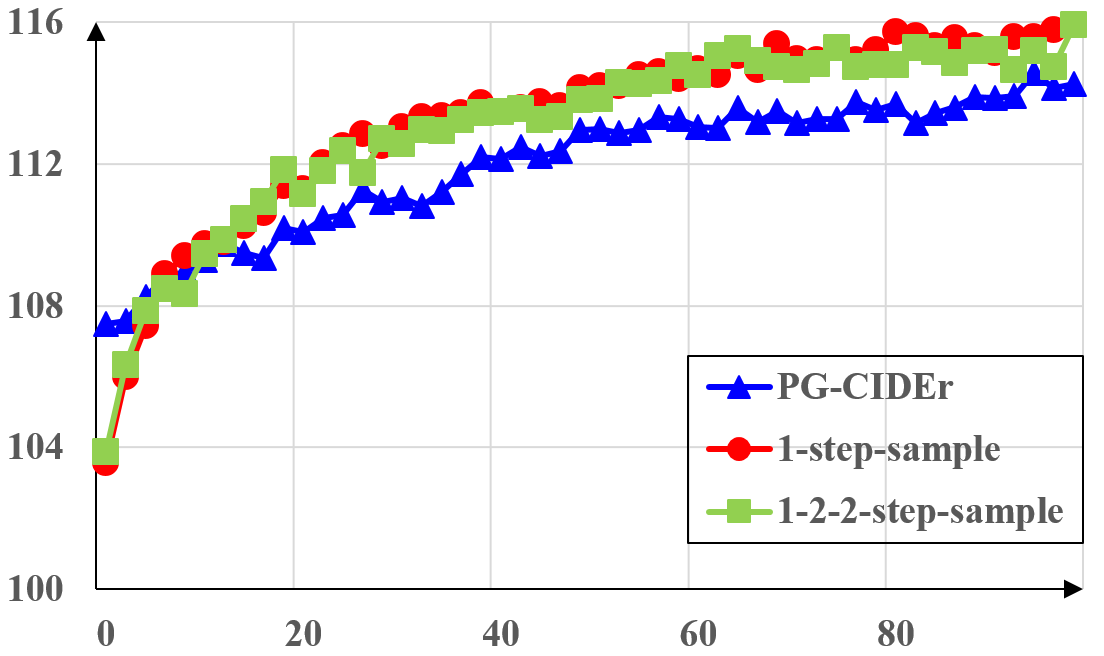}
				\end{minipage}
				\label{fig2:c}
			}&
			\subfigure[METEOR]{
				\begin{minipage}[]{1.5in}
					\centering
					\includegraphics[height=0.95in, width=1.6in]{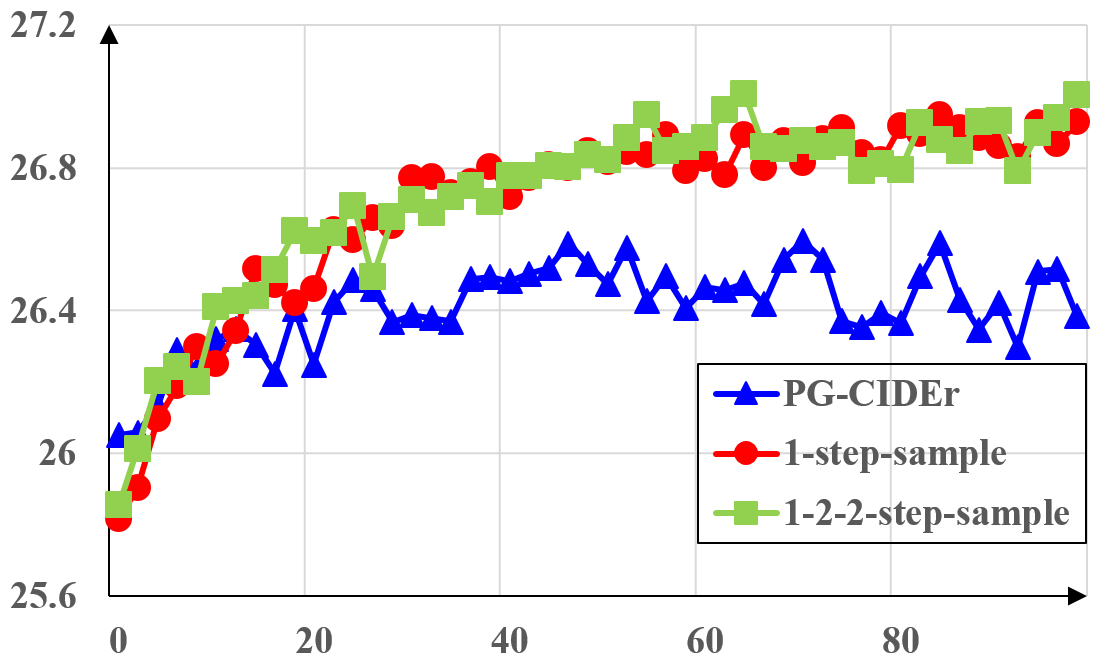}
				\end{minipage}
				\label{fig2:d}
			}
			\centering
		\end{tabular}
		\caption{(a)(b): Performance of SCST \cite{rennie2017self}, $1$-step-maxpro and $1$-$2$-$2$-step-maxpro; (c)(d): Performance of PG-CIDEr \cite{liu2017improved}, $1$-step-sample and $1$-$2$-$2$-step-sample. The horizotal axes are every $2K$ training steps and the vertical axes are corresponding metrics on validation set.}
		\label{fig2}
	\end{center}
\end{figure*}

\begin{figure*}[t]
	\begin{center}
		\begin{tabular}{cccc}
			\centering
			\subfigure[max-probability-mean]{
				\begin{minipage}[]{1.5in}
					\centering
					\includegraphics[height=0.95in, width=1.6in]{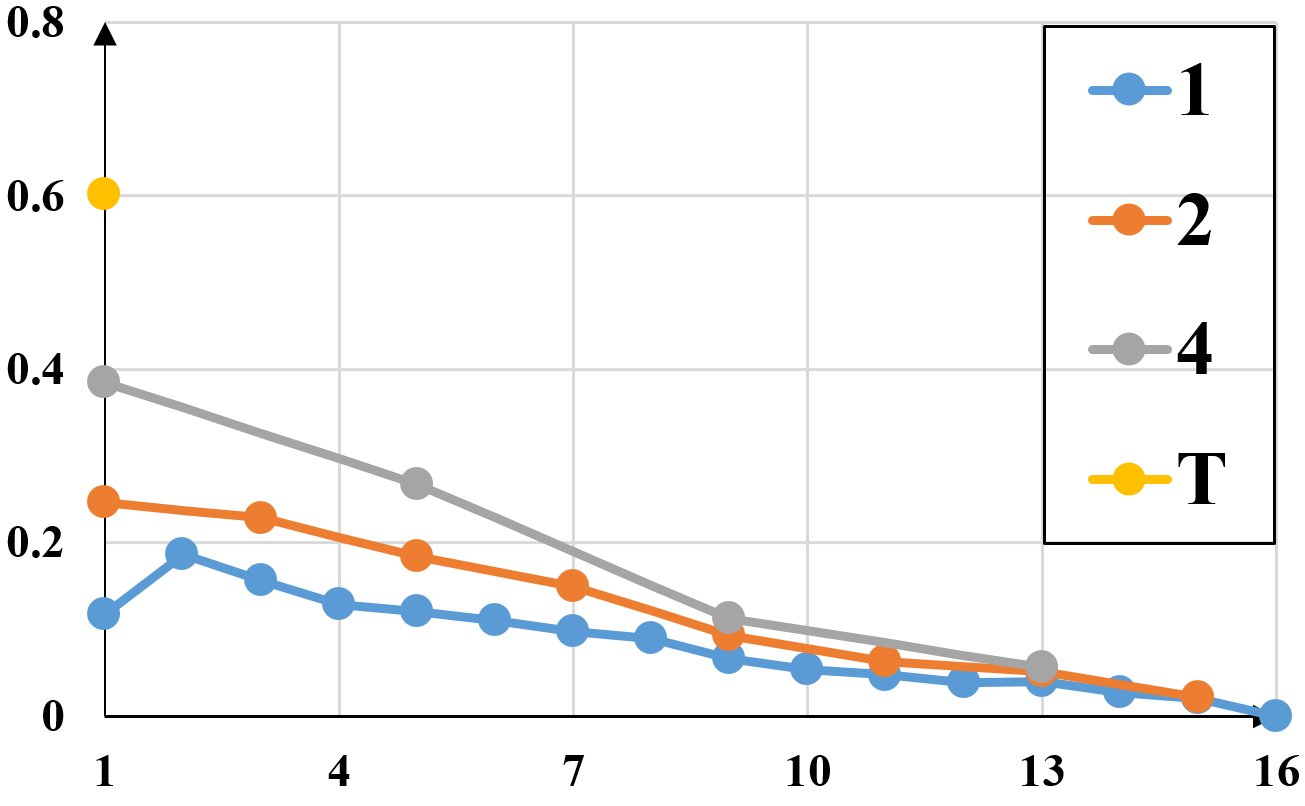}
				\end{minipage}	
				\label{fig2-5:a}
			}&
			\subfigure[max-probability-variance]{
				\begin{minipage}[]{1.5in}
					\centering
					\includegraphics[height=0.95in, width=1.6in]{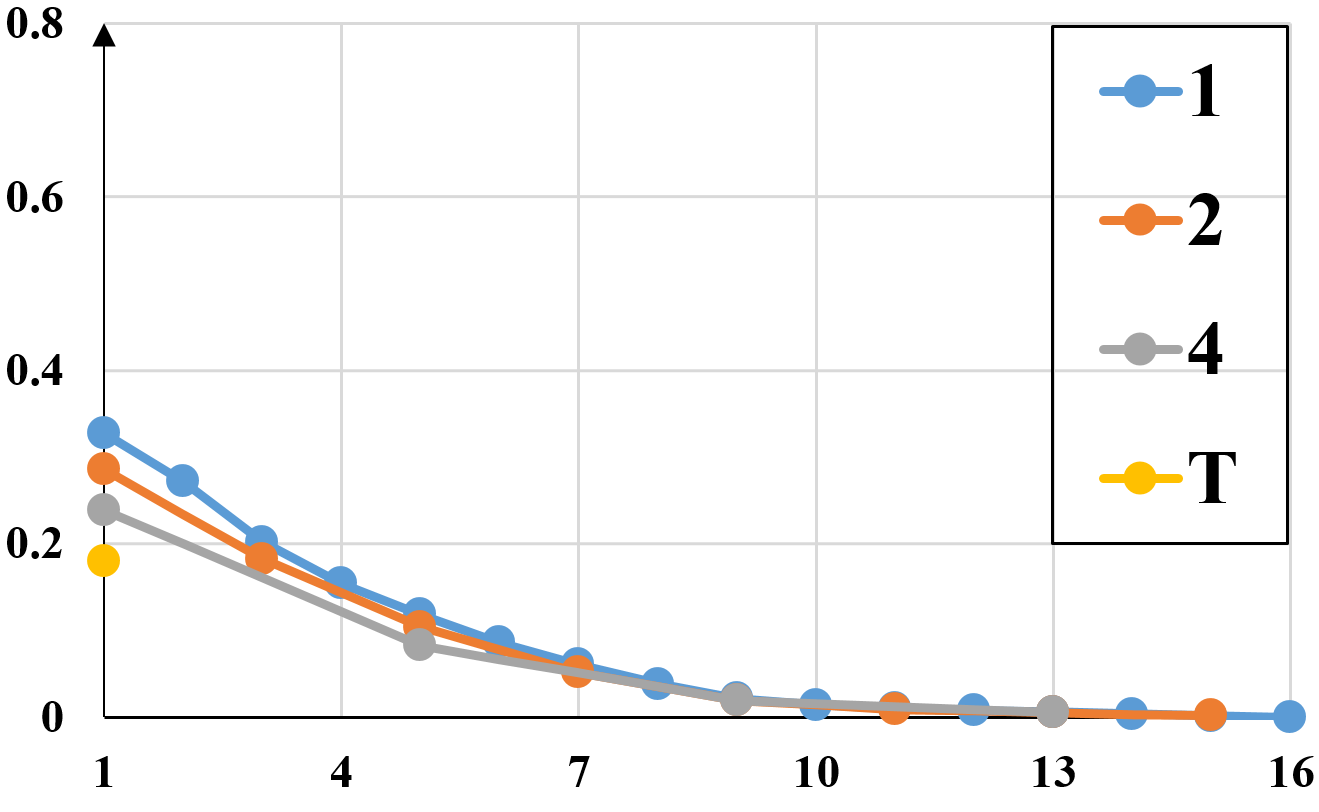}
				\end{minipage}
				\label{fig2-5:b}
			}&
			\subfigure[$K$-Monte-Carlo-mean]{
				\begin{minipage}[]{1.5in}
					\centering
					\includegraphics[height=0.95in, width=1.6in]{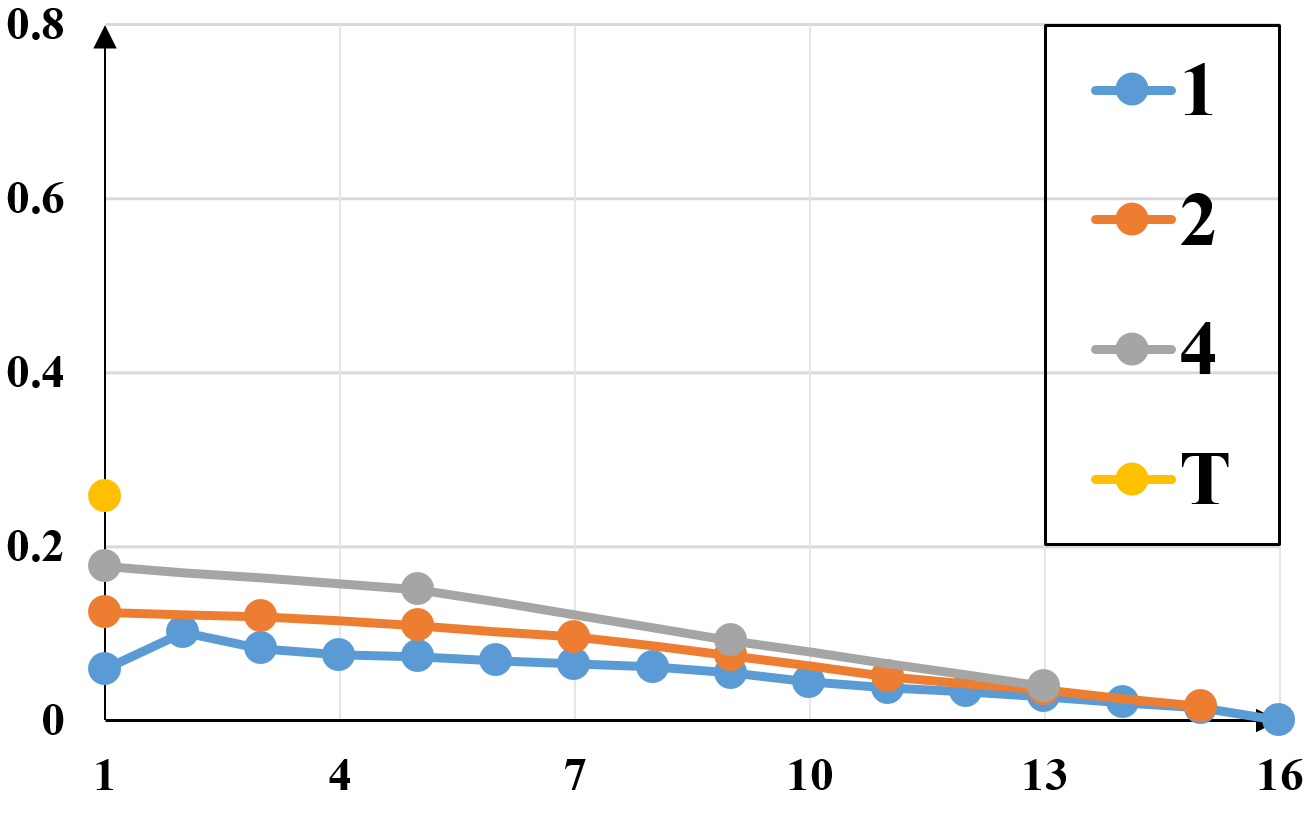}
				\end{minipage}
				\label{fig2-5:c}
			}&
			\subfigure[$K$-Monte-Carlo-variance]{
				\begin{minipage}[]{1.5in}
					\centering
					\includegraphics[height=0.95in, width=1.6in]{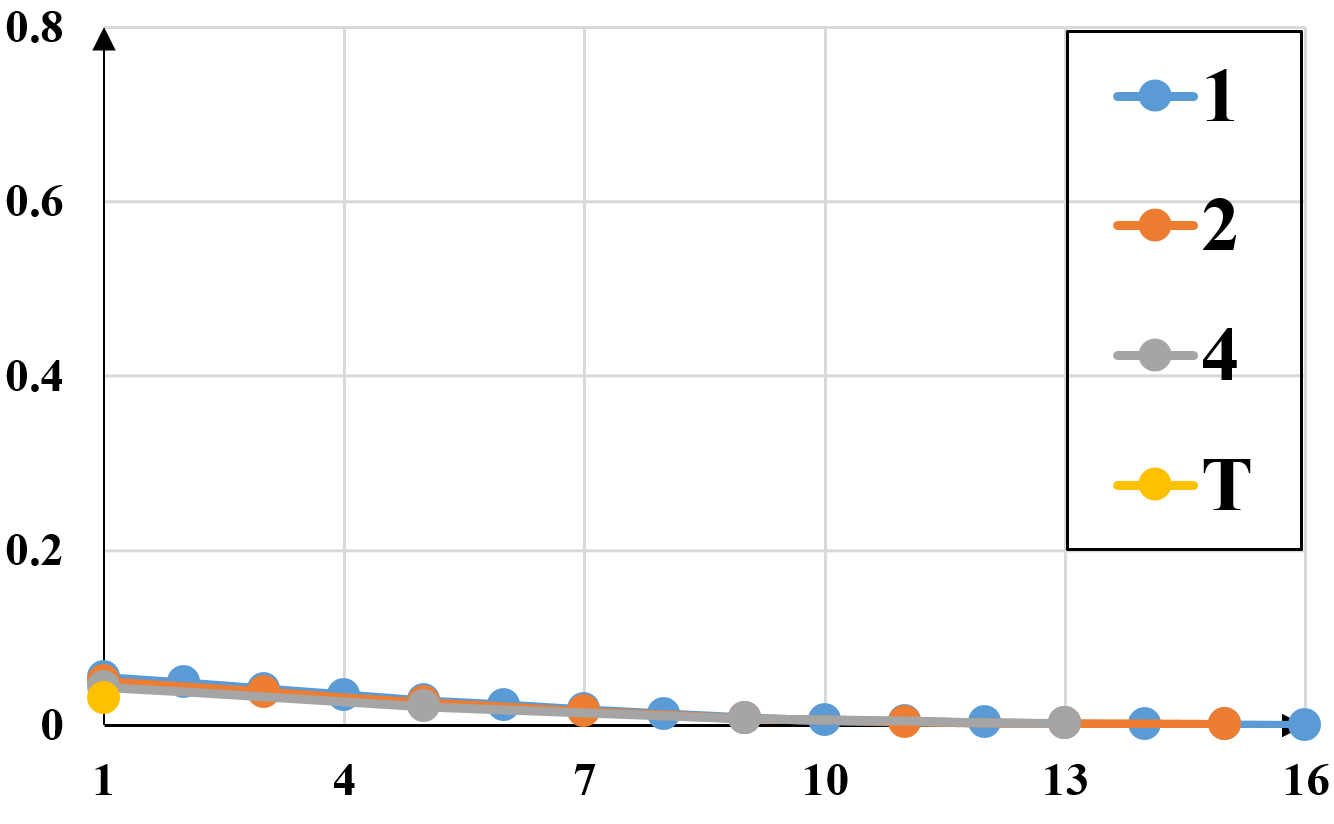}
				\end{minipage}
				\label{fig2-5:d}
			}
			\centering
		\end{tabular}
		\caption{Mean and variance of those $n$-step reformulated advantage in max-probability rollout (a)(b) and $K$ Monte Carlo rollouts (c)(d), where $n=\{1,2,4,T\}$. We do rollout $100$ times for each state-action pair after pretraining using cross entropy loss, calculate the mean and variance of reformulated advantage, and finally average all absolute value of the mean and variance of all training data in sequence time step order, where time step $t=\{1,2,...,16\}$ (horizotal axis).}
		\label{fig2-5}
	\end{center}
\end{figure*}

\begin{figure*}[!t]
	\begin{center}
		\begin{tabular}{ccc}
			\centering
			\subfigure[]{
				\begin{minipage}[t]{0.295\linewidth}
					\centering
					\includegraphics[height=1.26in, width=1.85in]{COCO_val2014_000000005032.jpg}
					{\scriptsize
						\leftline{a woman and a kid are standing on skis in the snow.}
						\leftline{\textcolor{red}{a woman and a child on skis in the snow.}}
						\leftline{\textcolor{blue}{a woman and a child standing on skis in the snow.}}
					}
				\end{minipage}
				\label{fig3:a}
			}&
			\subfigure[]{
				\begin{minipage}[t]{0.295\linewidth}
					\centering
					\includegraphics[height=1.26in, width=1.85in]{COCO_val2014_000000005754.jpg}
					{\scriptsize
						\leftline{a box of donuts of different colors and varieties.}
						\leftline{\textcolor{red}{a box of donuts and a variety of donuts.}}
						\leftline{\textcolor{blue}{a box of donuts sitting on top of a table.}}
					}
				\end{minipage}
				\label{fig3:b}
			}&
			\subfigure[]{
				\begin{minipage}[t]{0.295\linewidth}
					\centering
					\includegraphics[height=1.26in, width=1.85in]{COCO_val2014_000000009186.jpg}
					{\scriptsize
						\leftline{a wide variety of vases and chandelier in a window display.}
						\leftline{\textcolor{red}{a glass case with many different types of glass.}}
						\leftline{\textcolor{blue}{a display case with a bunch of vases on it.}}
					}
				\end{minipage}	
				\label{fig3:c}
			}\\
			\subfigure[]{
				\begin{minipage}[t]{0.295\linewidth}
					\centering
					\includegraphics[height=1.26in, width=1.85in]{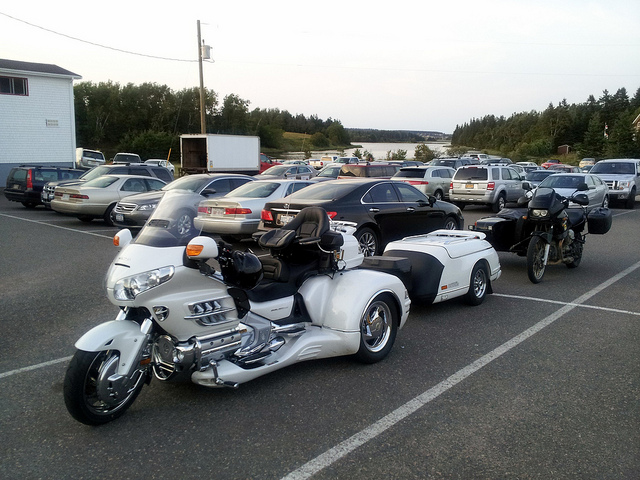}
					{\scriptsize
						\leftline{many cars and motorcycles are parked in a parking lot.}
						\leftline{\textcolor{red}{a motorcycle parked in a parking lot next to a parking lot.}}
						\leftline{\textcolor{blue}{a motorcycle parked in a parking lot with a group of cars.}}
					}
				\end{minipage}	
				\label{fig3:d}
			}&
			\subfigure[]{
				\begin{minipage}[t]{0.295\linewidth}
					\centering
					\includegraphics[height=1.26in, width=1.85in]{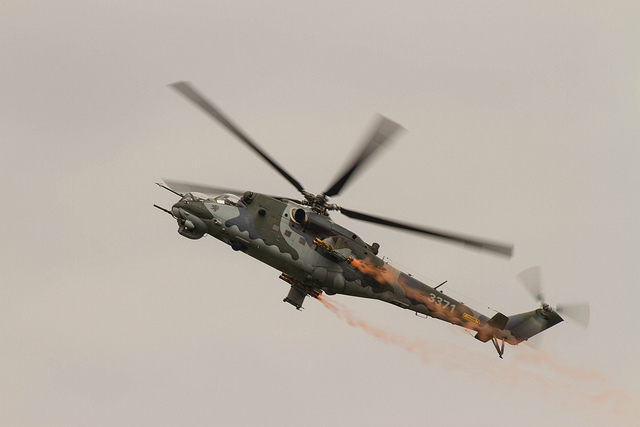}
					{\scriptsize	
						\leftline{a helicopter is flying upwards in the sky.}
						\leftline{\textcolor{red}{a black and white photo of a black and white photo.}}
						\leftline{\textcolor{blue}{a black and white photo of a helicopter flying in the sky.}}
					}
				\end{minipage}
				\label{fig3:e}
			}&
			\subfigure[]{
				\begin{minipage}[t]{0.295\linewidth}
					\centering
					\includegraphics[height=1.26in, width=1.85in]{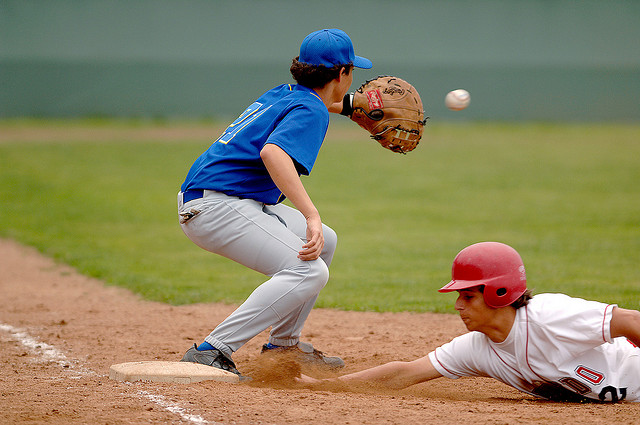}
					{\scriptsize
						\leftline{a man catching a baseball as another slides into the base}
						\leftline{\textcolor{red}{a baseball player is throwing a baseball}}
						\leftline{\textcolor{blue}{a baseball player throwing a ball on a field}}
					}
				\end{minipage}	
				\label{fig3:f}
			}
			\centering
		\end{tabular}
		\caption{Qualitative results of our model compared with Ground Truth and the model trained under XENT loss. Captions in black (first line), red(second line) and blue (third line) are ground truth captions, and those predicted by XENT and $1$-step-maxpro respectively.}
		\label{fig3}
	\end{center}
\end{figure*}

\subsection{Quantitative Analysis}
\noindent\textbf{Performance of the Karpathy test split.}
In Table \ref{tabel1}, we report the performance of our models, SCST\cite{rennie2017self} and PG-CIDEr \cite{liu2017improved} on the Karpathy test split, and all the models are single model. In general, we can see that our models have the best performance on all metrics. Comparing our basic model $1$-step-maxpro and $1$-step-sample with XENT, we obtain a significant improvement on CIDEr score over XENT at a great margin from $102.1\%$ to $115.1\%$ and $115.4\%$ of $1$-step-maxpro and $1$-step-sample respectively, since our basic models are reinforcement learning-based models and can address the exposure bias and directly optimize the evaluation metric. In particular, the $1$-step-sample outperform $1$-step-maxpro in terms of almost all metrics, and we can conclude that the average reward of $K$ Monte Carlo rollouts can estimate the more precise state-action value than max-probability rollout, which leads to better performance. However, $1$-step-sample need to sample $K$ rollouts with a greater computation cost. 

Regarding max-probability rollout, we compare different $n$-step-maxpro in Table \ref{tabel1}. We can see that intermediate settings $n=2,4$ attain better overall scores than two extremes $1$ and $T$ (SCST\cite{rennie2017self}). Better performance of intermediate settings originates from the fact that they increase the absolute value of the mean of reformulated advantage while lowering variance in most time steps compared to $n=1$, which are quantitatively shown in Fig.~\ref{fig2-5:a} \& \ref{fig2-5:b}. Since rollout-based methods estimate a rough state-action value, when $n=1$ reformulated advantage is small with large variance and it may turn out to be wrong to encourage or suppress a token in this strict greedy strategy. As $n$ increases, the dilemma will be eased but gradually loses per-token advantage until a sequence level advantage of $n=T$. This implies intermediate $n$ which balances the approximation of per-token advantage and the improvement of the absolute value of the mean of reformulated advantage, is always better in max-probability rollout. Moreover, different $n$ or combining different $n$ has different effects on balancing these two conflicts, \eg the performance of $n=2$ is better than that of $n=4$ and close to that of $1$-$2$-$2$, and $1$-$2$-$4$-$T$ and $T$-$4$-$2$-$1$ are both inferior to $1$-$2$-$2$. We also show the performance curves of the Karpathy validation split during training illustrated in Fig.~\ref{fig2}. In Fig.~\ref{fig2:a} \& \ref{fig2:b}, our models have an overwhelming advantage over SCST\cite{rennie2017self} throughout the whole training process, which demonstrates that self-critical per-token advantage is better than self-critical sequence level advantage.

Regarding $K$ Monte Carlo rollouts, $1$-step-sample and $1$-$2$-$2$-step-sample are superior to PG-CIDEr\cite{liu2017improved}, which demonstrates that self-critical per-token advantage is better than parametrized per-token advantage in Table \ref{tabel1} and Fig.~\ref{fig2:c} \& \ref{fig2:d}.

Comparing different effects of $n$-step towards max-probability rollout and $K$ Monte Carlo rollouts, we find that large $n$ can increase the absolute value of the mean of reformulated advantage while lowering the variance using these two kinds of rollouts in Fig.~\ref{fig2-5}. However, $1$-$2$-$2$-step-maxpro is superior to $1$-step-maxpro and $1$-$2$-$2$-step-sample is close to $1$-step-sample in Table~\ref{tabel1}. Therefore, $n$-step ($n=2$) is more effective in max-probability rollout than in $K$ Monte Carlo rollouts. It is possible because degrees of change in the absolute value of the mean and the variance of reformulated advantage across different $n$ are relatively small in $K$ Monte Carlo rollouts and thus possibly cannot offset the lose of per-token advantage, while those are relatively large in max-probability rollout and large $n$ (\eg. $n=2$) can balance better these two conflicts as illustrated in Fig.~\ref{fig2-5}.

\noindent\textbf{Performance on the official MSCOCO testing server.}
Table \ref{tabel2} shows the result of our single models and 4 ensembled model using beam search with beam size set to $3$ on the official MSCOCO evaluation server, and all other results are based on single model. Our single models and ensembled models outperform all of them in terms of most metrics, even the ones which use complex attention mechanisms \cite{you2016image,lu2017knowing}, and other reinforcement learning-based models which all introduce parameterized estimator \cite{ranzato2015sequence,liu2017improved,zhang2017actor} and sequence level advantage \cite{rennie2017self}. 

\subsection{Qualitative Analysis}
Fig.~\ref{fig3} shows some qualitative results of $1$-step-maxpro against Ground Truth and the model trained with XENT loss. Each image has three captions from these sources listed below. In general, the captions predicted by $1$-step-maxpro are better compared with the model trained with XENT loss. In Fig.~\ref{fig3:a}, we can see that when the image content is common in the dataset and not too complex to describe, XENT and $1$-step-maxpro can predict correct captions. Since the reinforcement learning-based model can avoid accumulating errors during generating the caption, the captions in Fig.~\ref{fig3:b}-\ref{fig3:e} generated by $1$-step-maxpro can describe more important objects and capture their relationships with more distinctive information of the image, while those generated by XENT are less descriptive or incorrect to some degree. When a variety of human activities that appear rarely in the dataset or different activities with the same objects that are difficult to distinguish by the model, the models easily have the incorrect prediction. For example, in Fig.~\ref{fig3:f}, $1$-step-maxpro and XENT both predict wrong captions that the player in the base is throwing the ball, who in fact is catching the ball with a glove.

\section{Conclusion}
We reformulate advantage function to estimate per-token advantage without using parametrized estimator. Moreover, $n$-step reformulated advantage is proposed to increase the absolute value of the mean of reformulated advantage while lowering variance. Our methods outperform state-of-the-art methods that use the sequence level advantage and parametrized estimator on MSCOCO benchmark.

\noindent\textbf{Acknowledgements}

This work was supported in part by the National Key R\&D Program of China (2017YFC0821005), National Basic Research Program of China (973 Program, 2015CB351800), and High-performance Computing Platform of Peking University, which are gratefully acknowledged.

{\small
\bibliographystyle{ieee}
\bibliography{egbib}
}

\end{document}